\documentclass[10pt,twocolumn,letterpaper]{article}

\usepackage{cvpr}
\usepackage{times}
\usepackage{epsfig}
\usepackage{graphicx}
\usepackage{amsmath}
\usepackage{amssymb}
\usepackage{multirow}
\usepackage{makecell}
\usepackage{color}

\usepackage[pagebackref=true,breaklinks=true,letterpaper=true,colorlinks,bookmarks=false]{hyperref}

\cvprfinalcopy 

\def\cvprPaperID{1329} 

\ifcvprfinal\fi
\begin{document}

\title{Video Playback Rate Perception for Self-supervised\\ Spatio-Temporal Representation Learning}

\author{Yuan Yao\textsuperscript{\rm 1}\footnotemark[1] , 
Chang Liu\textsuperscript{\rm 1}\footnotemark[1] , 
Dezhao Luo\textsuperscript{\rm 2}, Yu Zhou\textsuperscript{\rm 2} and Qixiang Ye\textsuperscript{\rm 1}\footnotemark[2] \\\\
\textsuperscript{1}{University of Chinese Academy of Sciences, Beijing, China}\\
\textsuperscript{2}{Institute of Information Engineering, Chinese Academy of Sciences, Beijing, China}\\
{\tt\small \{yaoyuan17,liuchang615\}@mails.ucas.ac.cn, \{luodezhao,zhouyu\}@iie.ac.cn}\\
{\tt\small qxye@ucas.ac.cn}
}
\maketitle
\renewcommand{\thefootnote}{\fnsymbol{footnote}}
\footnotetext[1]{Equal contribution} 
\footnotetext[2]{Corresponding author} 
\thispagestyle{empty}

\begin{abstract}
In self-supervised spatio-temporal representation learning, the temporal resolution and long-short term characteristics are not yet fully explored, which limits representation capabilities of learned models. In this paper, we propose a novel self-supervised method, referred to as video Playback Rate Perception (PRP), to learn spatio-temporal representation in a simple-yet-effective way. PRP roots in a dilated sampling strategy, which produces self-supervision signals about video playback rates for representation model learning. PRP is implemented with a feature encoder, a classification module, and a reconstructing decoder, to achieve spatio-temporal semantic retention in a collaborative discrimination-generation manner. The discriminative perception model follows a feature encoder to prefer perceiving low temporal resolution and long-term representation by classifying fast-forward rates.  The generative perception model acts as a feature decoder to focus on comprehending high temporal resolution and short-term representation by introducing a motion-attention mechanism.
PRP is applied on typical video target tasks including action recognition and video retrieval. Experiments show that PRP outperforms state-of-the-art self-supervised models with significant margins. Code is available at \href{https://github.com/yuanyao366/PRP}{\color{magenta}github.com/yuanyao366/PRP}.
\end{abstract}

\section{Introduction}

Deep networks, $i.e.,$ Convolutional Neural Networks (CNNs)~\cite{CNN2012}, have achieved unprecedented success in computer vision area. 
This can be largely attributed to the learned rich representation incorporating both low-level fine-details and high-level semantics~\cite{MELM2019}. 
To realize rich representation, networks are typically pre-trained using large-scale image/video datasets ($e.g.$, ImageNet~\cite{ImageNet16} and Kinetics ~\cite{kay2017kinetics}) under accurate annotation supervision~\cite{kim2019self}. 

However, large-scale data annotation is laborious, expensive, or can be impractical, particularly for complex data such as videos and concepts such as action analysis and video retrieval~\cite{fernando2017self,kay2017kinetics}. 
Considering the availability of large-scale unlabelled data on the Web, self-supervised representation learning, which leverages intrinsic correspondence within unlabelled data to pre-train desired representation models, has attracted increasing attention.  

\begin{figure}[t]
\begin{center}
\includegraphics[width=0.45\textwidth]{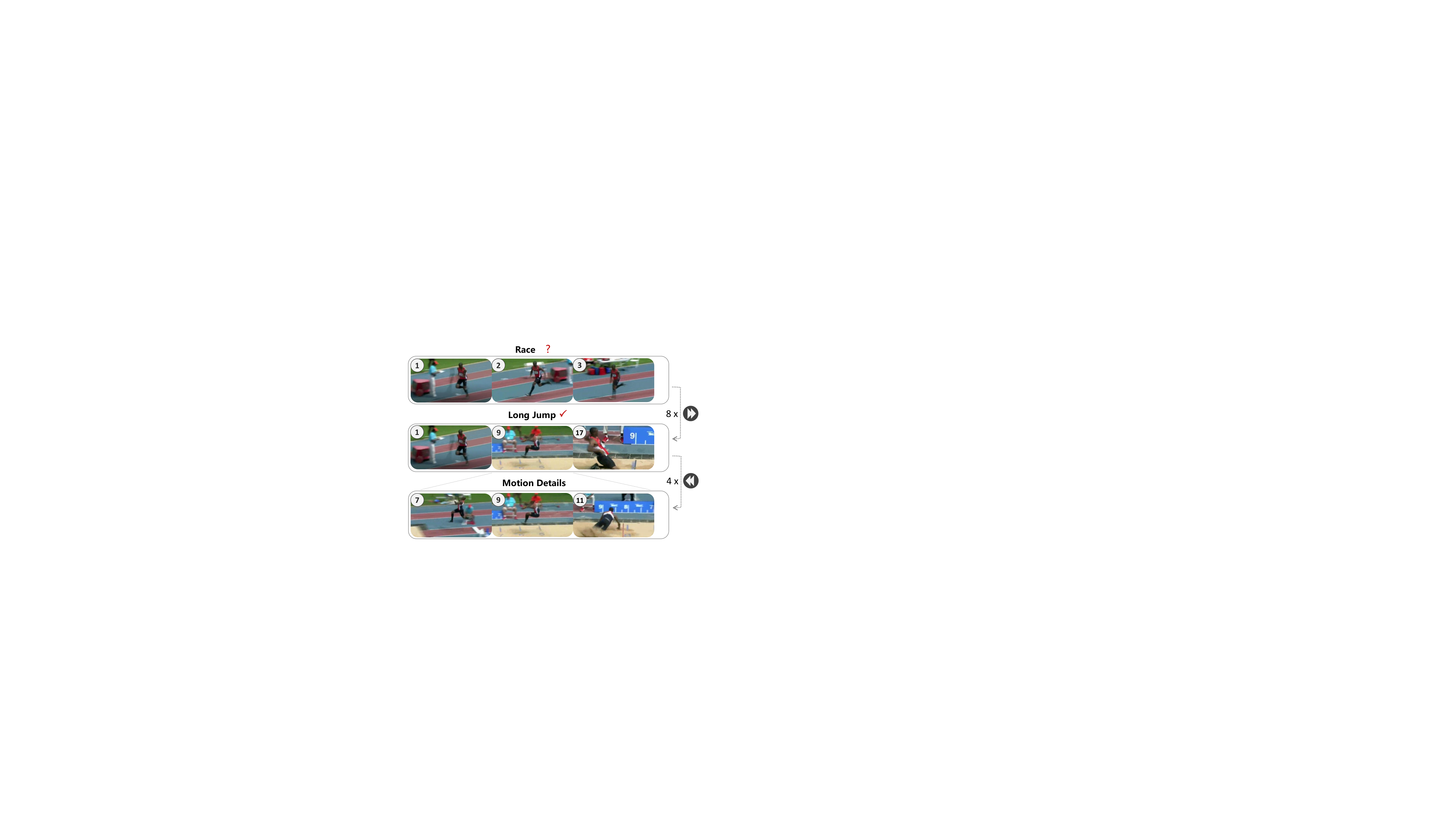}
\end{center}
   \caption{With limited visible frames, video clips with different playback rates (temporal resolutions) imply different semantics. A video clip with normal playback rate (first row) can be misunderstood as ``race". With higher playback rate (second row), we can see that it is in fact ``long jump", of which short-term motion details can be perceived in the slow-down video (third row). Perceiving videos with different playback rates is crucial in learning long-short term  spatio-temporal representation.}
\label{fig:long}
\end{figure}

Self-supervised representation learning defines an annotation-free proxy task, which leverages easily developed supervision signals from data itself to train network models, which then facilitate the implementation of the downstream target tasks.
From the perspective of frame content perception, early self-supervised methods focused on predicting the spatial transformation of images~\cite{fernando2017self}. Without considering the temporal relations, however, the learned features are merely on a frame-by-frame basis, which are inappropriate to video analysis tasks because the temporal dimension defines essential differences between a video sequence and an image set. Recent works~\cite{wang2019self} learned spatio-temporal representation by regressing both motion and appearance statistics. Nevertheless, without the capability to perceive temporal resolution characteristics, such a mechanism is unable to learn long-short term representation necessary for precise video understanding, Fig.\ \ref{fig:long}.


In this paper, we propose a novel self-supervised approach, referred to as video Playback Rate Perception (PRP), targeting at learning representation about multiple temporal resolutions in a simple-yet-effective manner. PRP is motivated by the motion perception mechanism observed in primate visual systems~\cite{Science1988,Neuron1994}, $i.e.$, different visual cells respond to different temporal changes. M-cells are sensitive to quick and short-term changes while P-Cells focus on slower and longer-term variation. This mechanism has been explored by SlowFast networks~\cite{SlowFast2019} for video recognition, while we update it to a self-supervised manner to perceive multiple temporal resolutions.

\begin{figure*}[t]
\begin{center}
\includegraphics[width=0.97\textwidth]{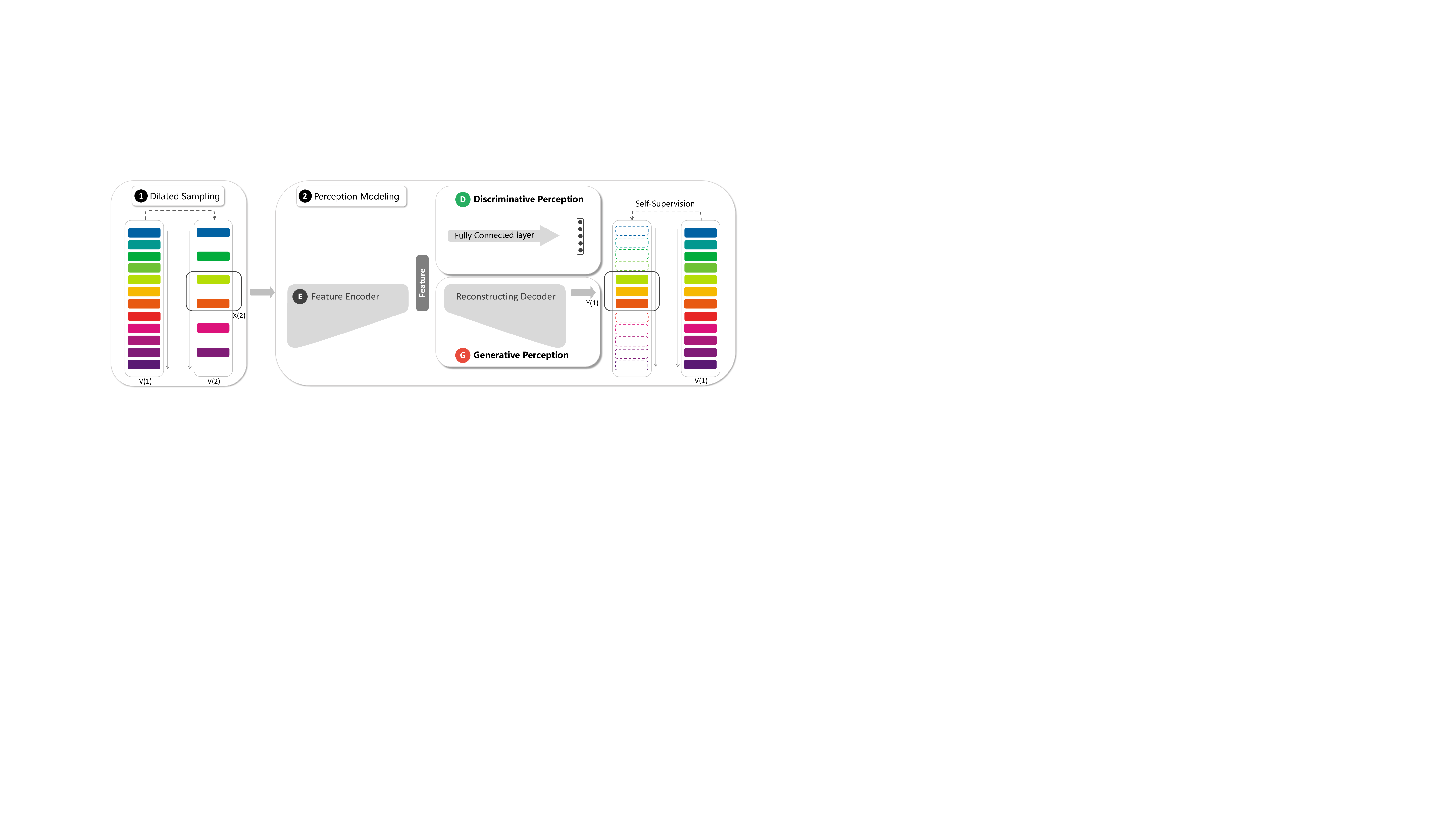}
\end{center}
   \caption{Playback rate perception (PRP) is composed of dilated sampling and perception modeling. Perception modeling is implemented with a feature encoder, a discriminative module, and a reconstructing decoder (generative module). The self-supervision signals are generated using dilated sampling.
   }
\label{fig:short}
\end{figure*}

To perceive temporal resolution characteristics within video data, a dilated sampling strategy is designed to produce videos with various playback rates. The original videos simulate high playback rates relative to frame-sampled videos, and content similarity between videos of different playback rates are used as a supervision signal for representation learning. 

With a discriminative model, PRP can be trained to classify videos of different playback rates. With a generative model, PRP is driven to reconstruct low playback rate videos from high playback rate ones. The discriminative perception model follows a feature encoder to focus on perceiving low temporal-resolution and long-term representation by classifying fast-forward rates. The generative perception model acts as a feature decoder to focus on comprehending high temporal-resolution and short-term representation by introducing a motion-attention mechanism. Collaborative discriminative-generative perception further aggregates long-short term representation capacity, Fig.\ \ref{fig:short}. 

The contributions of this work include:
\begin{itemize}
    \item A novel video Playback Rate Perception (PRP) approach is proposed to capture temporal resolution characteristics within video domain in a self-supervised manner.

    \item PRP is implemented with discriminative and generative perception models, which cooperatively retain spatio-temporal semantics in representation models. Furthermore, we introduce a motion attention mechanism, which drives representation to focus on meaningful foreground regions.

    \item We apply PRP to three kinds of 3D CNNs and two target tasks including action recognition and video retrieval, and improve the state-of-the-arts with significant margins.
    
\end{itemize}

\section{Related Work}

Self-supervised learning leverages information from unlabelled data to train models. Existing approaches usually define an annotation-free proxy task which demands a network predicting information hidden within unannotated videos.
The learned models can then be applied to target tasks (eitehr supervised or unsupervised) after fine-tuning. Conventional self-supervised methods
include discriminative proxy tasks such as classifying transformed images~\cite{gidaris2018unsupervised,kim2018learning,doersch2015unsupervised} or video content~\cite{zhao2017spatio}, and generative proxy tasks which include image inpainting ~\cite{pathak2016context} and video  reconstruction~\cite{vondrick2016generating,zhao2017spatio}.   

\subsection{Proxy Tasks}
From a broader view, proxy tasks can be constructed on top of multiple sensory data such as ego-motion~\cite{EgoMotion2017}, sound~\cite{AutoEncoding14}, and cross-modal data~\cite{LookListenLearn2016,LookListenLearn2017,SelfMultiModal2019}. Although in this paper, we mainly review proxy tasks based on visual signals.

\textbf{Spatial Representation Learning.} Spatial transforms applied to images can produce supervision signals for representation learning~\cite{larsson2017colorization}. As a representative method, the rotation-based self-supervised approach~ \cite{gidaris2018unsupervised,feng2019self} learns CNN features by rotating images and using rotated angles as supervision. The completion-based approach~ \cite{kim2018learning,doersch2015unsupervised,SelfBenchmark2019} learns image representations by predicting damaged Jigsaw puzzles. While context impainting~\cite{Inpainting2016} trains the CNN model to predict content of a withheld image region conditioned according to its surroundings, the image-patch matching approach ~\cite{Invariance2017,FreeAnchor2019} trains a representation model to capture spatial in-variance. 

\textbf{Spatio-temporal Representation Learning.} The large amount of video clips with rich spatio-temporal information provide various supervision signals. In ~\cite{WangXiaoLong2015}, the temporal continuity of video frames could be used as a supervisory signal. In ~\cite{misra2016shuffle,lee2017unsupervised}, predicting orders of frames or video clips drives learning spatio-temporal representation. In~\cite{fernando2017self}, an odd-one-out network was proposed to identify the unrelated or odd clips from a set of otherwise related clips. To find the odd clip, the models have to learn spatio-temporal features which can discriminate similar clips. In~\cite{WatchingMove2017}, unsupervised motion segmentation on videos was used to obtain segments, which perform as pseudo ground truth to train CNNs for segmentation.

Early methods usually learn features based upon 2D CNNs and simplistically based on a frame-by-frame process, which are inappropriate to video analytic tasks where spatio-temporal features are prevailing.
Recently, 3D representations are learned~\cite{wang2019self} by regressing motion and appearance statistics. The order of video clips is then used as a supervised signal for temporal representation learning ~\cite{xu2019self}. 3D CNN models are trained by completing space-time cubic puzzles~ \cite{kim2019self}.

Despite of substantial progress in the field, existing methods unfortunately ignore the multiple temporal resolutions, which are essential for video-based tasks. Without these temporal resolution characteristics, the representation capability of learned models remains limited.

\subsection{Target Tasks} 

For video-related tasks 3D CNN models were trained using a large-scale video databases with video category annotation~\cite{feichtenhofer2016convolutional,tran2015learning}. Nevertheless, the representation models trained on video classification tasks lack general applicability. Fine-tuning such models to other target tasks, $e.g.$, action recognition and video retrieval, could produce sub-optimal results. To conquer these issues, we propose the self-supervised PRP approach, and target at improving the model generality, by incorporating long-short term temporal representations, 

\section{Playback Rate Perception}

Fast-forward and slow-down playback are two commonly used modes when browsing videos. To quickly understand video content, $e.g.,$ a movie, we can use the fast-forward mode. To capture the fine details within a wonderful clip, we usually require action replay with a slow-down play rate. The way humans perceive video content demonstrates an important fact that the temporal resolution and long-short term characteristics are critical to get better understanding of videos. 

Based on this observation, we propose the video Playback Rate Perception (PRP) for representation learning, which is composed of two components: dilated sampling and perception modeling. Dilated sampling augments video clips into different temporal resolution (fast-forward) while perception modeling learns rich spatio-temporal representation to classify videos into playback rates and/or reconstruct from the low temporal resolution videos to high temporal resolution ones (slow-down), Fig.\ \ref{fig:short}.

\subsection{Dilated Sampling}

Given a raw video $V(1)$, we uniformly sample a video frame from each $s$ frames with the same temporal interval, which is denoted as $s \times$ dilated sampling. This procedure generates video $V(s)$ with $s \times$ fast-forward playback rate. Considering the spatial similarity and temporal ambiguity among video frames, we sample successive $l$ frames from $V(s)$ as a learning sample, $X(s)$, which can be fed to 3D CNNs. For the example shown in Fig.\ \ref{fig:short}(left), $s=2$ and $l=2$. The videos $V(s)$ with different dilated sampling intervals have consistent content but different playback rates. Such playback rates, together with their corresponding video content, provide self-supervision signals for representation model learning.

\begin{figure}[t]
\begin{center}
\includegraphics[width=0.4\textwidth]{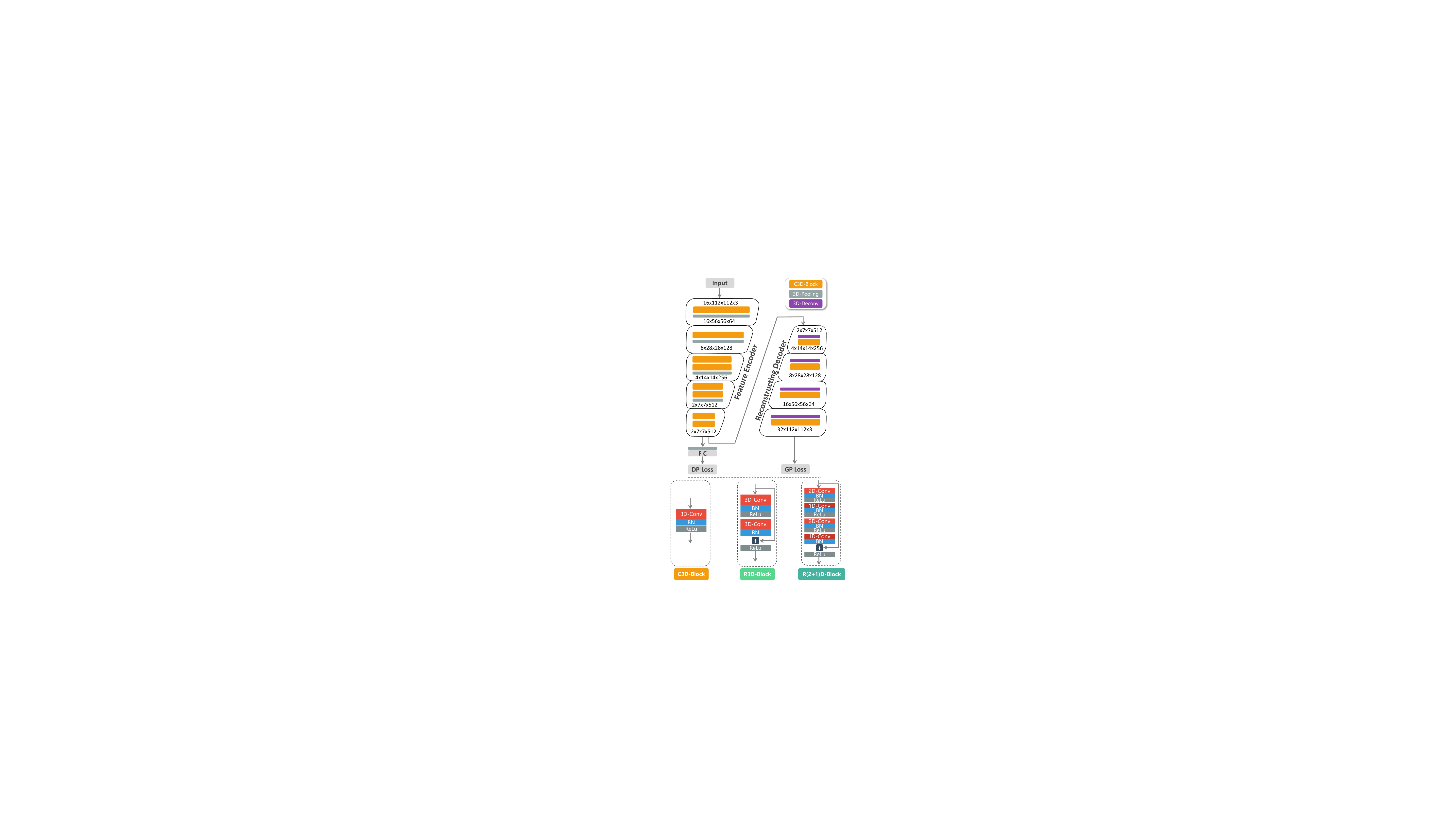}
\end{center}
   \caption{Up: encoder-decoder structure. Down: C3D, R3D, and R(2+1)D blocks.}
\label{fig:network}
\end{figure}

\subsection{Perception Modeling}

\textbf{Feature Encoder.}
To extract both spatial and temporal features, we choose C3D \cite{tran2015learning}, R3D and R(2+1)D \cite{tran2018closer} as feature encoders. C3D is a natural extension from 2D CNNs for spatio-temporal representation learning as it can model the temporal information of videos. It stacks five C3D blocks which consist of a classic 3D convolution with the kernel size of $t \times k \times k$ followed by a batch normalization layer and a ReLU layer. As shown in Fig.\ \ref{fig:network}, we take C3D backbone as an example to build the feature encoder and show the dimensional transformation of each block.

R3D refers to 3D CNNs with residual connections. As shown in Fig.\ \ref{fig:network}, R3D block consists of two 3D convolution followed by batch normalization and ReLU layers. The input and output are connected with a residual unit before the last ReLU layer. In R(2+1)D, the overall structure is similar tp R3D. The 3D convolution is decomposed into a spatial 2D convolution and a temporal 1D convolution with additional batch normalization and ReLU layers attached.

\textbf{Discriminative Perception.}
As shown in Fig.\ \ref{fig:network}, features of the input video clip extracted by the encoder is fed to a classification model to predict the playback rate. The ground-truth label is denoted as $s_{c}$, where $1 \leq c \leq C$, $C$ is the number of different sampling intervals of the inputs. This procedure can be referred to as discriminative perception upon a normalized probability $p_{c}$ of which the input video clip belongs to class $c$, $p_{c} = \frac{\exp(a_{c})}{\sum_{c=1}^C \exp(a_{c})}$, where $a_{c}$ is the $c$-${th}$ output of the fully connected layer. Based on the normalized probability, the parameter $\theta$ for the network model is updated by optimizing a cross entropy loss, as
\begin{equation}   
   \arg\min_{\theta}\mathcal{L}_{d}= -\sum _{c}^C s_c\log p_c.
  \label{eq:dploss}
\end{equation}

To optimize Eq.\ \ref{eq:dploss}, the feature encoder is driven to perceive subtle differences of motion intensity and scenario dynamics among adjacent frames which is essential for precise spatio-temporal representation. 

\textbf{Generative Perception.}
Beyond discriminative perception we further propose a generative perception mode to promote PRP's understanding capacity, which targets at reconstructing the $r \times$ slow-down video clips. The reconstruction procedure is performed with a feature decoder network which has four 3D deconvolutional blocks, Fig.\ \ref{fig:network}. For each decoder block, we stack a deconvolutional layer with stride $2\times2\times2$ followed by a C3D block. To generate a video with reconstructing rate $r$ ($r$ times as slow as the input video), the fourth deconvolutional takes a stride of $r\times2\times2$.

\textit{Ground-Truth.}
To predict the interpolated frames, we set the dilated sampling interval as $s = 2^{k_{1}}, (k_{1} = 0, 1, 2, \cdots)$ and the reconstructing rate as $r = 2^{k_{2}}, k_{2} \in {0, 1, 2} $. The ground-truth of the input clip $X(2^{k_{1}})$ with $2^{k_{2}}\times$ slow-down generation can be sampled from the video $V(2^{k_{1} - k{2}})$. As shown in Fig.\ \ref{fig:short}(right), a $2\times$ slow-down generative perception is implemented by taking the $2\times$ dilated sampled video clip as input and the raw video as output (self-supervision signal). If $k_{2} > k_{1}$, we can use linear interpolation to generate the ground-truth clip from the raw video.

\begin{figure*}[t]
\begin{center}
\includegraphics[width=0.95\textwidth]{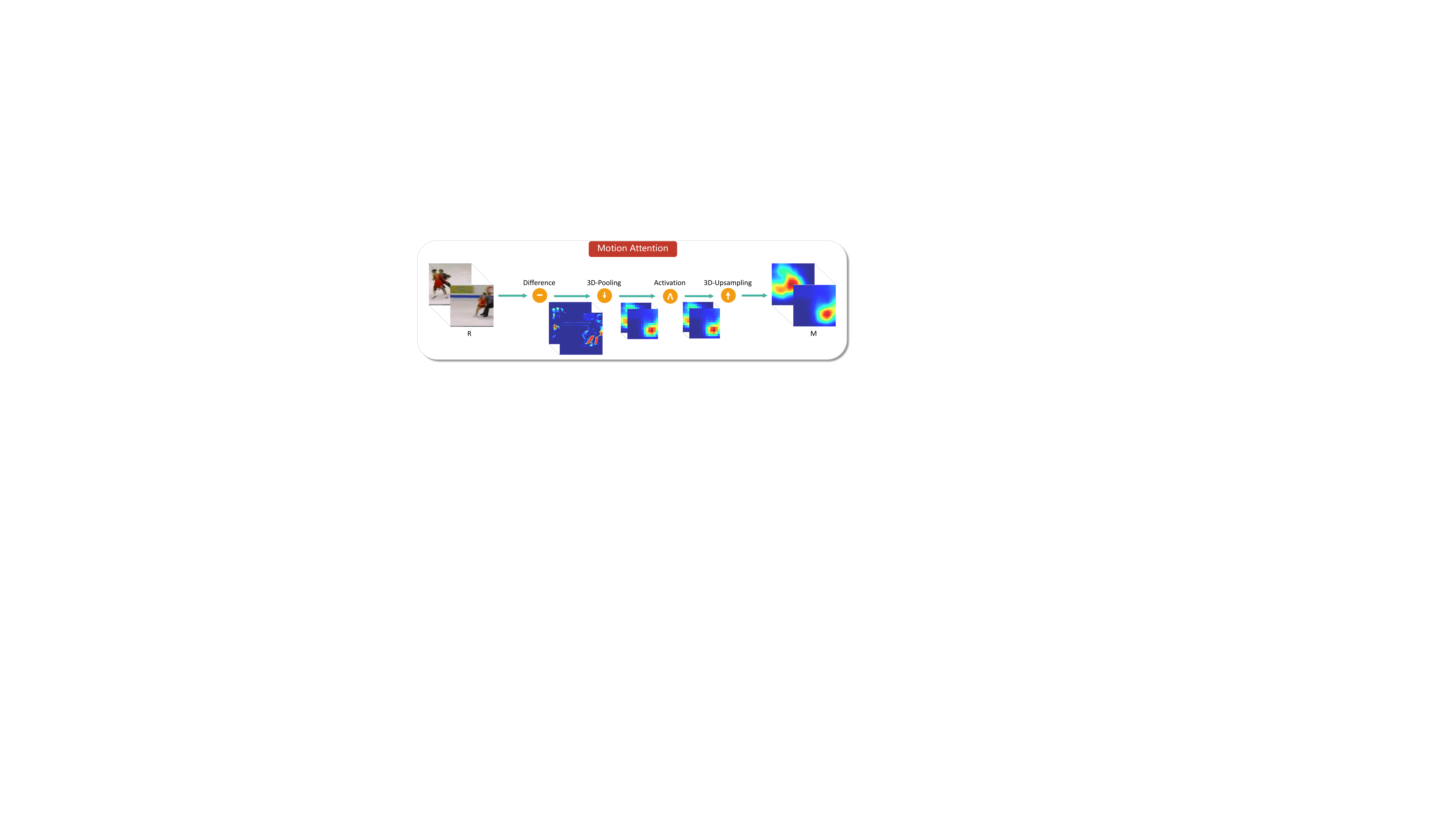}
\end{center}
   \caption{Calculation of motion attention based on frame difference, 3D-Pooling, activation and 3D-Upsampling operations. }
\label{fig:MA}
\end{figure*}

\textit{Motion Attention.} To reconstruct video clips, MSE~\cite{hasan2016learning} loss is commonly used to build a generative network. It is important to note that our PRP is not designed to generate high quality videos but to learn long-short term video representations. To fulfill this purpose, we propose a motion attention regularized MSE (m-MSE) loss, which drives the network concentrating on reconstructing and interpolating frame regions in significant motion. 

Denoting the $t$-$th$ ground-truth frame for slow-down generation, the $t$-$th$ motion attention map and the $t$-$th$ predicted video frame as $G^t=(g^t_{ij})$,  $M^t = (m_{ij}^{t})$ and $Y^t=(y^t_{ij})$,  the m-MSE loss can be defined as 
\begin{equation}
    \arg\min_\theta\mathcal{L}_{g}=\frac{1}{N}\sum_{t, i, j}{m_{ij}^{t}(y_{ij}^{t}-g_{ij}^{t})^2} ,
\end{equation}     
where $N$ is the number of pixels in the predicted video clip. $(ij)$ denotes a spatial location on video frames.

As shown in  Fig.\ \ref{fig:MA}, the motion attention maps $M$ are calculated according to the raw video frames $X(1)$ (denoted as $R$) which is an $s\times$ slow-down video clip of input $X(s)$, and through four steps including difference, 3D-Pooling, activation and 3D-Upsampling. 
In the difference step, adjacent frames $R^{t}$ and $R^{t+1}$ from the raw video clip are used to calculate the $t$-$th$ frame difference map $D^{t}$ as $D^{t} = \mathcal{D}(R^{t}, R^{t+1}) = |R^{t} - R^{t+1}|^{2}$. Considering that the frame difference maps can be affected by accidental noise as well as missing static foregrounds, a 3D-Pooling operation $\mathcal{P}$, as a spatio-temporal filer, is conducted on the difference maps to make it more consistent with foregrounds and more stable in the spatio-temporal domain. Then, an increasing activation function $\mathcal{A}$ is used to transform the pixel value of the difference maps to [$\lambda_{1}$, $\lambda_{2}$], $0 \leq \lambda_{1} \leq 1$ and $1 \leq \lambda_{2}$. Finally, a 3D-Upsampling operation $\mathcal{U}$ is applied to obtain motion attention maps of the same size with the ground-truth video frames.
The overall process of motion attention map generation is formulated as

\begin{equation}
    M = \mathcal{M}({R}) = \mathcal{U}(\mathcal{A}(\mathcal{P}(\mathcal{D}(R)))).
    \label{eq:motion-attention}
\end{equation}

\textbf{Discriminative-Generative Perception.}
To further learn richer spatio-temporal representations, discriminative and generative perception models are fused, Fig.\ \ref{fig:short}, by optimizing the following objective function, as
\begin{equation}   
  \arg\min_{\theta}\lambda_{d}\mathcal{L}_{d}+\lambda_{g}\mathcal{L}_{g}.
  \label{eq:dp}
\end{equation}
Fusion is performed in a cooperative manner, as the classification model is good at identifying long-term representation for playback rate discrimination, while the generative model can capture short-term fine-details for content reconstruction. With end-to-end learning, Fig.\ \ref{fig:short}, spatio-temporal characteristics of multiple temporal resolution can be encoded within the model.

\begin{figure}[t]
\begin{center}
\includegraphics[width=0.42\textwidth]{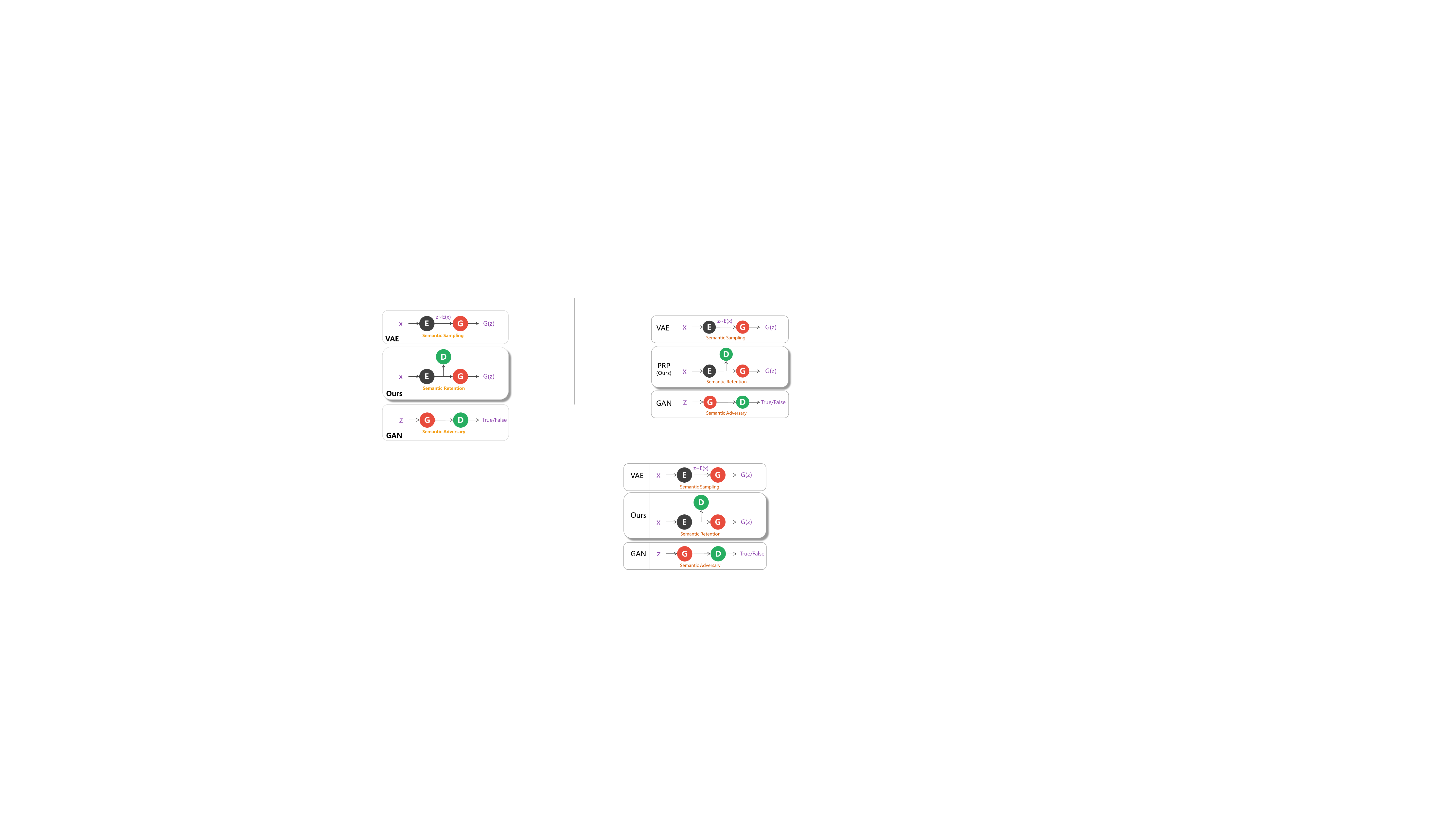}
\end{center}
   \caption{Comparison of Variational Auto-Encoder (VAE), Generative Adversarial Network (GAN), and the proposed encoder-decoder model. `E', `D', and `G' denote `Encoder', `Discriminator'  and `Generator', respectively.}
\label{fig:discussion}
\end{figure}

\subsection{Discussion} 

The proposed encoder-decoder framework contributes a new feature learning strategy, which is neither identical to Variational Auto-Encoder (VAE)~\cite{VAE2014} nor to Generative Adversarial Network (GAN)~\cite{GAN2019}, Fig.\ \ref{fig:discussion}. Specifically, our framework is driven by discriminative and generative models to achieve semantic retention, which means that the encoded temporal semantics can be transferred to downstream target tasks, as much as possible. By contrast, VAE targets at semantic sampling controlled by the latent variable ($z$) following normal distribution. The encoder in VAE should learn features that best represent the distribution of inputs while the generator uses specified features for data generation conditioned on the latent variable. 

Like GAN, our approach involves both generative and discriminative models. The essential difference is that GAN leverages models in an adversarial manner while ours works cooperatively. GAN uses the generative model to produce images which are difficult to be classified by the discriminative model. Our approach learns general semantics, $i.e.,$ multi-resolution spatio-temporal representation, in a cooperative discrimination-generation manner.

\section{Experiments}
We first elaborate experimental settings for PRP, and then evaluate various sampling intervals and reconstructing rates with ablation study on a target task (action recognition).  We then analyze how PRP drives the model focusing on foreground regions and perceiving long-short term spatio-temporal characteristics. Finally, we evaluate the performance of PRP by applying the self-supervised models on target tasks including video action recognition and video retrieval, and compare it with state-of-the-art methods.

\subsection{Experimental Setting}
\textbf{Datasets.} Two action recognition datasets, UCF101 \cite{soomro2012ucf101} and HMDB51 \cite{jhuang2011large}, are used to demonstrate the effectiveness of PRP. UCF101 is collected from websites including Prelinger archive, YouTube and Google videos, containing 101 action categories with 9.5k videos for training and 3.5k videos for testing. HMDB51 is extracted from a variety of sources ranging from digitized movies to YouTube. It is consists of 51 action categories with 3.4k videos for training and 1.4k videos for testing. Both datasets exhibit challenges include intra-class variance of actions, complex camera motions, and cluttered backgrounds. To perform action recognition and retrieval on these datasets requires learning rich spatio-temporal representation.

\textbf{Network Architecture.} 
In video encoder, C3D, R3D, R(2+1)D are used as network backbones, where the kernel size of 3D convolutional layers is set to $3\times 3\times 3$. 
In video generation, four deconvolutional layers are stacked and followed by C3D blocks. To generate a video which is $r$ times as slow as the input video, we set the $4$-$th$ deconvolutional layer with a stride of $r\times2\times2$, where the reconstructing rate $r$ is determined through ablation study.

\textbf{Motion Activation.} 
To calculate motion attention maps, the activation function $\mathcal{A}$ in Eq.\ \ref{eq:motion-attention} is implemented as
$\mathcal{A}(D) = \frac{\lambda_{2}-\lambda_{1}}{max(D)-min(D)}{(D-min(D))}+{\lambda_{1}}$, where $D$ is the frame difference map. $\lambda_{1}$ is empirically set to 0.8 and $\lambda_{2}$ 2.0. 
We use an 3D-AveragePooling with kernel size $15 \times 28 \times 28$ and stride size $16 \times 7 \times 7$.
The 3D-Upsampling operation is set to tri-linear mode.

\textbf{Parameters.} 
Following the settings in ~\cite{tran2015learning, tran2018closer}, we set the length of input video $X(s)$ $l=16$ and determine the dilated sampling interval $s\in S$ through ablation study. 
During training, we randomly split 800 videos from the training set as validation set. Video frames are resized to $128\times171$ and randomly cropped to $112\times112$ as data augmentation. 
We empirically set the parameters $\lambda_{d}$, $\lambda_{g}$ for loss balance as 0.1 and 1.
With a initial learning rate 0.01, momentum 0.9 and weight decay 0.0005, the pre-training process is carried out for 300 epochs. 
The learned representation model with the lowest validation loss is used for target tasks.

\subsection{Ablation study}
In this section, we conduct experiments on the first split of UCF101 to analyze the effect of PRP under different dilated sampling intervals, different reconstructing rates, with/without motion attention.

\begin{table}
\begin{center}
\begin{tabular}{l c c}
\hline
Samp. Interval & Random acc.(\%) & DP acc. (\%)\\
\hline
\{1,2\} & 50 & 88.3 \\
\{1,2,4\} & 33 & 80.1 \\
\{1,2,4,8\} & 25 & 69.7 \\
\{1,2,4,8,16\} & 20 & 60.1 \\
\hline
\end{tabular}
\end{center}
\caption{Classification accuracy of the discriminative perception (DP) model under different sampling intervals.}
\label{table:DP acc}
\end{table}

\begin{table}
\begin{center}
\begin{tabular}{c|l c c}
\hline
Method & \makecell[c]{Samp. Interval} & Rec. Rate & UCF101(\%)\\
\hline
Random & \makecell[c]{-} & - & 62.0\\
\hline
\multirow{4}{*}{DP} 
& \{1,2\} & - & 68.3 \\
& \{1,2,4\} & - & 68.7 \\
& \{1,2,4,8\} & - & \textbf{69.9}\\
& \{1,2,4,8,16\} & - & 67.9\\
\hline
\multirow{4}{*}{GP} & \{1,2,4,8\} &1 (w/o \textit{MA}) & 67.1 \\
& \{1,2,4,8\} & 1 (w/ \textit{MA}) & 68.1 \\
& \{1,2,4,8\} & 2 (w/ \textit{MA}) & 68.2 \\
& \{1,2,4,8\} & 4 (w/ \textit{MA}) & \textbf{68.4} \\
\hline
DG-P & \{1,2,4,8\} & 2 (w/ \textit{MA}) & \textbf{70.9}\\
\hline
\end{tabular}
\end{center}
\caption{Ablation study of different model perception methods with corresponding different model parameters. The figures refer to action recognition accuracy on UCF101. ``Sam.Rate" and ``Rec.Rate" respectively denote sampling interval and reconstructing rate. ``DP", ``GP", and ``DG-P" respectively denote discriminative perception, generative perception, and discriminative-generative perception. ``\textit{MA}" denotes \textit{Motion Attention}. }
\label{table:ablation acc}
\end{table}

\textbf{Dilated sampling interval.} 
As shown in Table \ref{table:DP acc}, discriminative perception accuracy is consistently higher than random accuracy, which indicates that the discriminative perception model can learn effective spatio-temporal representation. 
Specifically, as the sampling interval $s$ increases, discriminative perception accuracy gradually decreases from 88.3\% to 60.1\%, while the accuracy of the target task increases from 68.3\% of \{1,2\} to 69.9\% with sampling interval \{1,2,4,8\}, Table~\ref{table:ablation acc}. It manifests that to some extent, larger sampling intervals force the model perceiving longer motion information which improves the representation capability of the learned model. However, when it comes to \{1,2,4,8,16\}, the video content jumps too much to be well perceived which makes the model confuse to learn discriminative representations. Therefore, the action recognition accuracy stops increasing. We thus set a sampling interval $s\in\{1,2,4,8\}$ in the following experiments. 


\textbf{Reconstructing rate.} 
%
As shown in Table \ref{table:ablation acc}, with reconstructing rate $r$ increasing, the performance increases from 68.1\% to 68.4\% when motion attention loss is applied, which can be explained that large reconstruction rate $r$ can force the network focusing on motion details, which is helpful for video understanding.
Considering the performance of $r = 2$ is comparable to which of $r = 4$, we set $r = 2$ as default in what follows to reduce the computational cost of the network.

\textbf{Discriminative and Generative Perception.}
As shown in Table \ref{table:ablation acc}, discriminative perception improves the action recognition accuracy from 62.0\% to 69.9\%, while generative perception improves the accuracy from 62.0\% to 68.4\%. 
The discriminative-generative model further improves the accuracy to 70.9\%, which validates the effectiveness of cooperative work of these two branches.

\textbf{Motion Attention.} The motion attention mechnism can drives representation to focus on meaningful foreground regions. As shown in Table \ref{table:ablation acc}, the application of motion attention boosts the accuracy from 67.1\% to 68.1\%, which is also a significant margin considering the challenging action recognition task.

\subsection{Visualizing Self-supervised Representation}

\begin{figure}[t]
     \centering
     \includegraphics[width=0.48\textwidth]{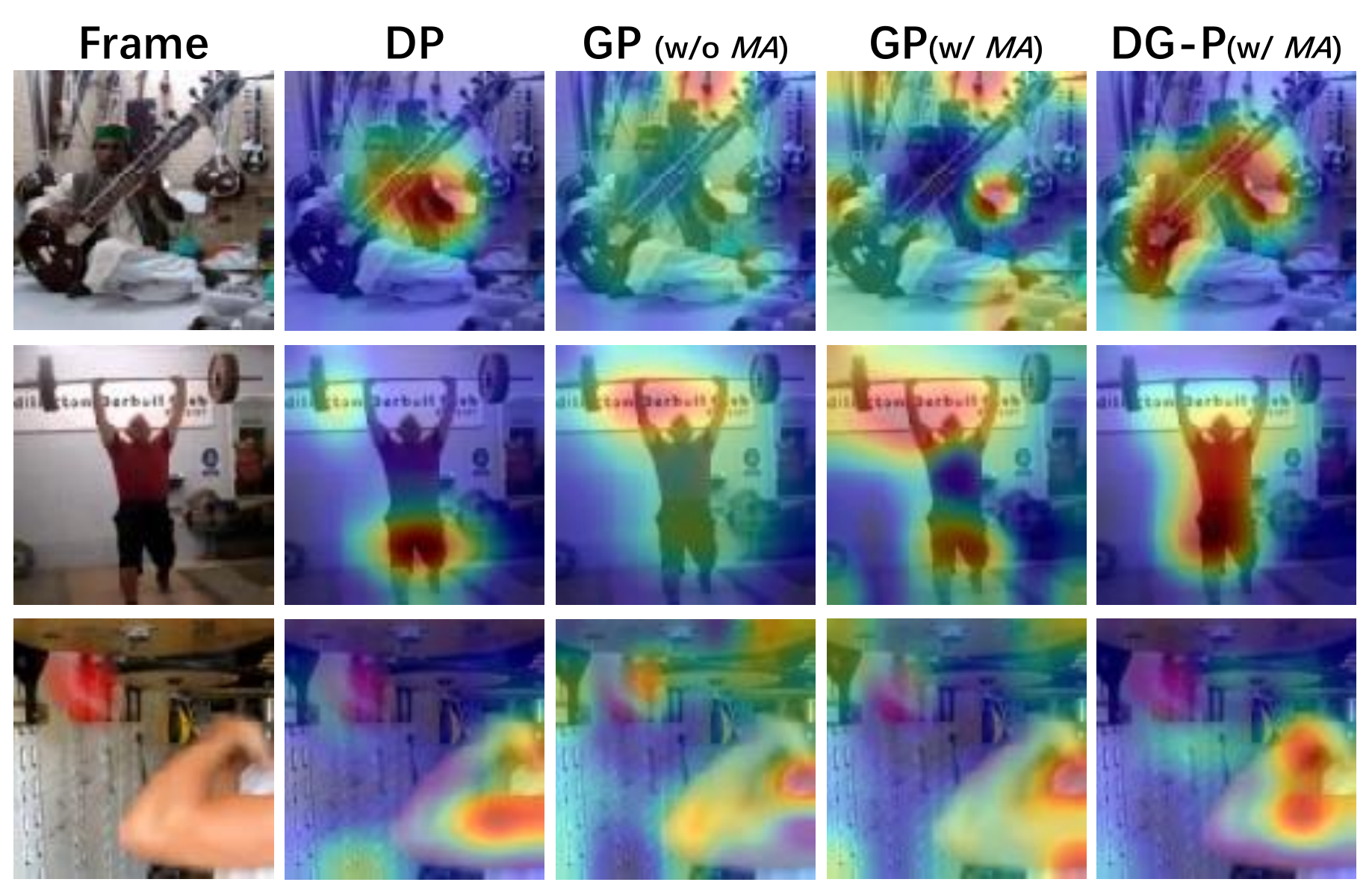}
     \caption{Visualization of activation maps.The attention maps are generated by summarizing convolutional feature  channels in conv5 layer ~\cite{AttentionTransfer2017} .``DP", ``GP", and ``DG-P" respectively denote discriminative perception, generative perception, and discriminative-generative perception. ``\textit{MA}" denotes \textit{Motion Attention}. }
     \label{fig:attention map}
\end{figure}

We try to understand what PRP learns by visualizing the feature activation maps, which indicating where the spatio-temporal representation focuses on. In Fig.\ \ref{fig:attention map}, we visualize and compare different perception models' activation maps on video frames. It can be seen that the discriminative perception model (DP) learns features sensitive to incomplete foreground regions containing major motion information, while the generative perception (GP) model learns features sensitive to where motion occurs but diverse to more context regions. With motion attention preferring to enhance motion areas, the generative perception model produces activation map with more motion areas activated. By fusing these two models, the learned features focus on complete foreground regions, which implies that the representation model incorporates long-short term motion information.

\begin{table}
    \centering

    \begin{tabular}{lcc}
    \hline
    Method&UCF101(\%)&HMDB51(\%)  \\
    \hline
    Jigsaw\cite{noroozi2016unsupervised}&51.5&22.5 \\

    OPN\cite{lee2017unsupervised} &56.3&22.1\\
    B\"uchler\cite{buchler2018improving}  &58.6&25.0 \\
    Mas\cite{wang2019self} & 58.8&32.6\\
    3D ST-puzzle\cite{kim2019self}& 65.0 &31.3\\
    ImageNet pre-trained&67.1&28.5\\
    \hline
    C3D(random) &61.8&24.7\\
    C3D(VCOP\cite{xu2019self}) &65.6& 28.4\\ 
    C3D(PRP) &\textbf{69.1}& \textbf{34.5}\\ 
    \hline
    R3D(random) &54.5&23.4\\
    R3D (VCOP\cite{xu2019self}) &64.9& 29.5\\ 
    R3D (PRP) &\textbf{66.5}& \textbf{29.7}\\     
    \hline
    R(2+1)D(random) &55.8&22.0\\
    R(2+1)D(VCOP\cite{xu2019self}) &\textbf{72.4}& 30.9\\ 
    R(2+1)D(PRP) &72.1& \textbf{35.0}\\
    \hline
    \end{tabular}
    
    \vspace{0.1cm}
    \caption{Performance comparison of self-supervised methods for spatio-temporal representation learning on UCF101 and HMDB51.} 
    \label{fig:state of the art}
\end{table}

\begin{figure*}[t]
\begin{center}

\includegraphics[width=0.95\textwidth]{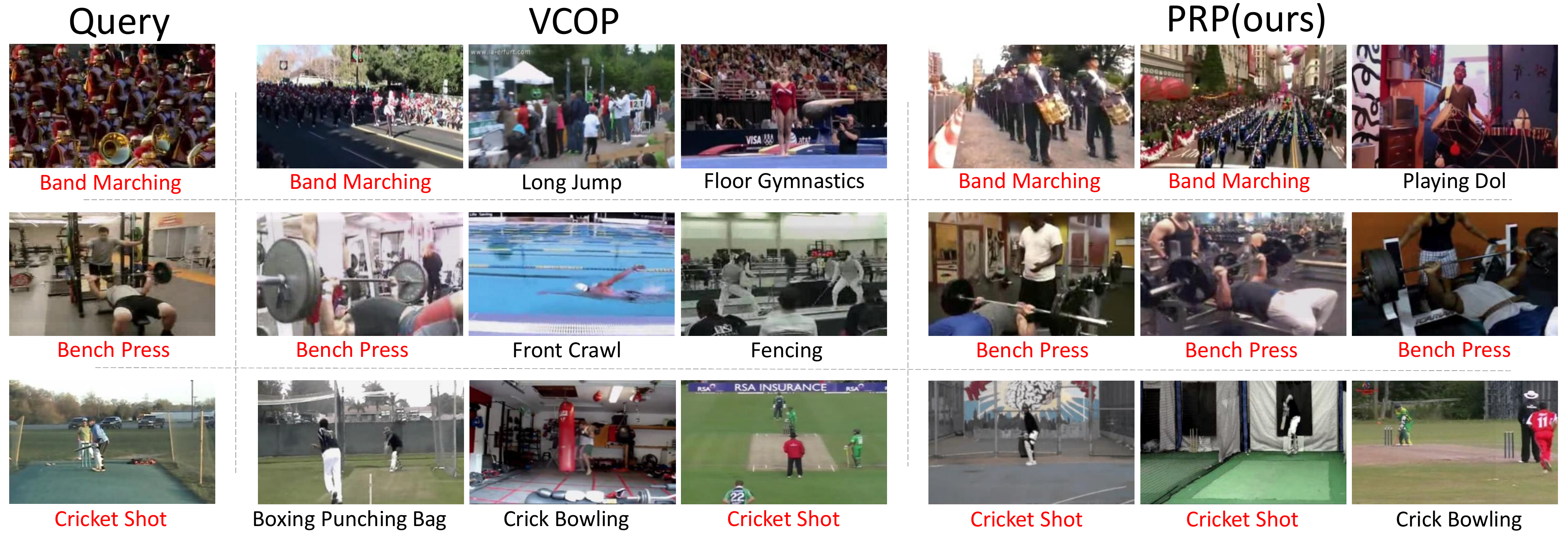}
   \caption{Comparison of video retrieval results. Red fonts indicate correct retrieval results. It can be seen that compared with the state-of-the-art VCOP approach, PRP achieves more accurate and reasonable video retrieval results. (Best viewed in color)}
\label{fig:retri}
\end{center}
\end{figure*}

\subsection{Evaluating Self-supervised Representation}
\textbf{Action Recognition.}  To verify our findings, we conduct experiments on action recognition which is a representative target task to validate the effectiveness of self-supervised representation~\cite{xu2019self}. 
For action recognition, we initialize the backbones with the model pre-trained on the first split of UCF101 by PRP, and fine-tune on UCF101 and HMDB51, Table \ref{fig:state of the art}. Data pre-processing and experimental settings are the same as those during PRP training. We feed features extracted by the backbones to fully-connected layers and obtain the category prediction.
For training, the fine-tuning procedure stops after 150 epochs. For testing, we follow the protocol of \cite{tran2018closer} and sample 10 clips for each video. The predictions on the sampled clips are then averaged to obtain the final prediction results. And we average classification accuracy over 3 splits for fair comparison. 

With the C3D backbone, our PRP approach obtains 69.1\% and 34.5\% which is 7.3\% and 9.8\% better than random initialization on UCF101 and HMDB51 respectively, Table \ref{fig:state of the art}.
Our PRP approach also obtains 3.5\% and 6.1\% better results compared with state-of-the-art VCOP approach \cite{xu2019self}, which are significant margins for the challenging action recognition task. 
With the R(2+1) backbone, PRP achieves 16.3\% (72.1\% vs. 55.8\%) and 13.0\% (35.0\% vs. 22.0\%) improvement over the random initialization. Our PRP approach also outperforms VCOP with significant margins. With the obtained results, we validate that PRP is able to learn richer spatio-temporal representations of videos compared with previous methods.

\textbf{Video Retrieval.}
To further verify its effectiveness, PRP is tested on the target task of nearest-neighbor video retrieval. As the video retrieval task is conducted with features extracted by the backbone network without fine-tuning, they largely rely upon the representative capacity of self-supervised model. An experiment is carried out on the first split of UCF101, following the protocol in \cite{xu2019self}. In the process of retrieval, video convolutional features are extracted with the backbone pre-trained by PRP. Each video in the test set is used to query $k$ nearest videos from the training set based upon their spatio-temporal features. When the category in the retrieved result is identical to that in the test video, we count this as the correct retrieval.

In Table \ref{fig:retrieve ucf101} and Table \ref{fig:retrieval hmdb}, we show top-1, top-5, top-10, top-20, and top-50 retrieval accuracy, which shows that PRP outperforms the state-of-the-art method equivalent on all evaluation metrics by substantial margins ( 8.7$\sim$10.7\% for top1 accuracy on UCF101). In Fig.\ \ref{fig:retri}, qualitative results further show PRP's superiority. 

\begin{table}
    \resizebox{1.0\columnwidth}{!}{
    \centering
    \begin{tabular}{lccccc}
    \hline
    Methods&top1&top5&top10&top20&top50  \\
    
        \hline
    Jigsaw\cite{noroozi2016unsupervised} &19.7&28.5&33.5&40.0&49.4 \\
    OPN\cite{lee2017unsupervised} &19.9&28.7&34.0&40.6&51.6 \\
    B\"uchler\cite{buchler2018improving} &25.7&36.2&42.2&49.2&59.5 \\
        \hline
    C3D(random) &16.7&27.5&33.7&41.4&53.0 \\
    
    C3D(VCOP\cite{xu2019self}) & 12.5 & 29.0&39.0&50.6&66.9\\

    C3D(PRP)&\textbf{23.2}&\textbf{38.1}&\textbf{46.0}&\textbf{55.7}&\textbf{68.4} \\
    \hline
    R3D(random) &9.9&18.9&26.0&35.5&51.9 \\
    R3D(VCOP\cite{xu2019self}) & 14.1 & 30.3&40.4&51.1&66.5\\

    R3D(PRP)&     \textbf{22.8}&\textbf{38.5}&\textbf{46.7}&\textbf{55.2}&\textbf{69.1} \\

    \hline
    R(2+1)D(random) &10.6&20.7&27.4&37.4&53.1 \\
   R(2+1)D(VCOP\cite{xu2019self}) & 10.7 & 25.9&35.4&47.3&63.9\\

     R(2+1)D(PRP)& 
         \textbf{ 20.3}&\textbf{34.0}&\textbf{41.9}&\textbf{51.7}&\textbf{64.2} \\

    \hline
    \end{tabular}
    }
    \vspace{0.1cm}
    \caption{Video retrieval performance on UCF101.}
    \label{fig:retrieve ucf101}
\end{table}

\begin{table}
    \resizebox{1.0\columnwidth}{!}{
    \centering
    \begin{tabular}{lccccc}
    \hline
    Methods&top1&top5&top10&top20&top50  \\
        \hline
    C3D(random) &7.4&20.5&31.9&44.5&66.3 \\
    C3D(VCOP\cite{xu2019self}) & 7.4 & 22.6&34.4&48.5&70.1\\
    C3D(PRP) &\textbf{10.5}&\textbf{27.2}&\textbf{40.4}&\textbf{56.2}&\textbf{75.9}\\

    \hline
    R3D(random) &6.7&18.3&28.3&43.1&67.9 \\
    R3D(VCOP\cite{xu2019self}) & 7.6 & 22.9&34.4&48.8&68.9\\
    R3D(PRP) & \textbf{8.2}&\textbf{25.8}&\textbf{38.5}&\textbf{53.3}&\textbf{75.9} \\

    \hline
            
    R(2+1)D(random) &4.5&14.8&23.4&38.9&63.0 \\
    R(2+1)D(VCOP\cite{xu2019self}) & 5.7 & 19.5&30.7&45.8&67.0\\

    R(2+1)D(PRP)&\textbf{8.2}&\textbf{25.3}&\textbf{36.2}&\textbf{51.0}&\textbf{73.0} \\
    \hline
    \end{tabular}
    }
    \vspace{0.1cm}
    \caption{Video retrieval performance on HMDB51.}
    \label{fig:retrieval hmdb}
\end{table}

\section{Conclusion}
In this paper, we proposed a novel video Playback Rate Perception (PRP) approach for self-supervised spatio-temporal representation learning. 
With a simple dilated sampling strategy, we augmented videos into different temporal-resolutions, which were then used to learn the long-short term characteristics of videos with discriminative and generative models.
Self-supervised models were applied on video action recognition and video retrieval tasks.
Extensive experiments showed that self-supervised models, trained with PRP, outperformed state-of-the-art self-supervised models with significant margins. 
Our work presented a promising direction and a new framework for self-supervised spatio-temporal representation learning.

\section*{Acknowledgement}
This work was supported in part by the National Natural Science Foundation of China (NSFC) under Grant 61836012, 61671427 and 61771447, and the National Key R\&D Program of China (2017YFB1002400).
{\small
\bibliographystyle{ieee_fullname}
\bibliography{main.bbl}
}
\end{document}